\PassOptionsToPackage{table}{xcolor}
\RequirePackage{fix-cm}
\documentclass{style}

\usepackage[utf8]{inputenc}
\usepackage{amsmath,amssymb}
\usepackage{enumitem,float,wrapfig,nicefrac}
\usepackage{needspace}
\usepackage{pgfplots}
\pgfplotsset{compat=1.18}
\newcommand{\nameofmethod}{DC-SAE}

\title{DC-SAE: Deep Compression Semantic Autoencoder for Faster Diffusion Convergence}
\author{%
  Xu Huang\texorpdfstring{$^{1*}$}{}\hspace{0.14cm}
  Ye Huang\texorpdfstring{$^{1*}$}{}\hspace{0.14cm}
  Zijun Liao\texorpdfstring{$^{1*}$}{}\hspace{0.14cm}
  Yuwei Niu\texorpdfstring{$^{1}$}{}\hspace{0.14cm}
  Xiaojie Li\\[0.35em]
  Menghan Zhou\texorpdfstring{$^{2}$}{}\hspace{0.14cm}
  De Wen Soh\texorpdfstring{$^{2}$}{}\hspace{0.14cm}
  Xiaotong Li\texorpdfstring{$^{1}$}{}\hspace{0.14cm}
  Daquan Zhou\texorpdfstring{$^{1\dagger}$}{}
}
\affiliation[1]{Peking University}
\affiliation[2]{Singapore University of Technology and Design}
\contribution[*]{Equal contribution}
\contribution[\dagger]{Corresponding author}
\abstract{
High-compression tokenizers are essential for scaling latent image generative models. However, aggressive compression creates a fundamental tradeoff between reconstruction fidelity and generation efficiency: high compression image encoder always increases the learning difficulty of diffusion training, resulting in slow model convergence.
Recent representation autoencoders speed up the diffusion training by improving the latent feature's expressive capability by replacing VAE encoders with pretrained semantic encoders, yet they are typically limited to moderate compression and lose pixel-level details necessary for faithful reconstruction. 
To achieve both high compression and fast diffusion training, we propose DC-SAE, a Decoupled Compact Semantic Autoencoder designed for high-compression image generation with accelerated diffusion model convergence.
DC-SAE consists of two key components: (1) a macro-level architecture design that leverages semantic encoders to enable higher compression ratios, and (2) a pixel-level encoder that preserves low-level details, ensuring high-fidelity image reconstruction.
We empirically demonstrate that DC-SAE performs strongly on image generation tasks, achieving both compact latent representations and efficient training dynamics.
Specifically, on the ImageNet dataset with $512 \times 512$ resolution, DC-SAE achieves  $32\times$ spatial compression, with \textbf{29.79} PSNR and \textbf{3.37} gFID, substantially outperforming the previous state-of-the-art high-compression tokenizer baselines DC-AE by 13.5\% and 54.9\% on PSNR and gFID, respectively, maintaining comparable throughput and faster diffusion model training convergence.
Beyond class-conditional generation, a $1.6$B-parameter DiT using DC-SAE achieves \textbf{0.84} on GenEval and \textbf{86.007} on DPG-Bench for text-to-image generation at $1024\times1024$ resolution.
}

\newcommand{\linkicon}[1]{\raisebox{-0.22em}{\includegraphics[height=1.15em]{resources/#1}}\hspace{0.35em}}
\makeatletter
\renewcommand\checkdata[2][]{\addtolist[#1]{#2}{\checkdatalist}{\checkdataformat}{\par\vskip 2.2mm}}
\makeatother
\checkdata[\linkicon{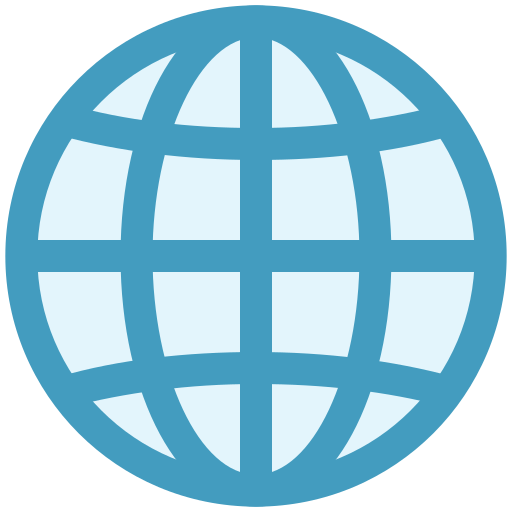}Project Page]{\href{https://dagroup-pku.github.io/DCSAE}{\texttt{https://dagroup-pku.github.io/DCSAE}}}
\checkdata[\linkicon{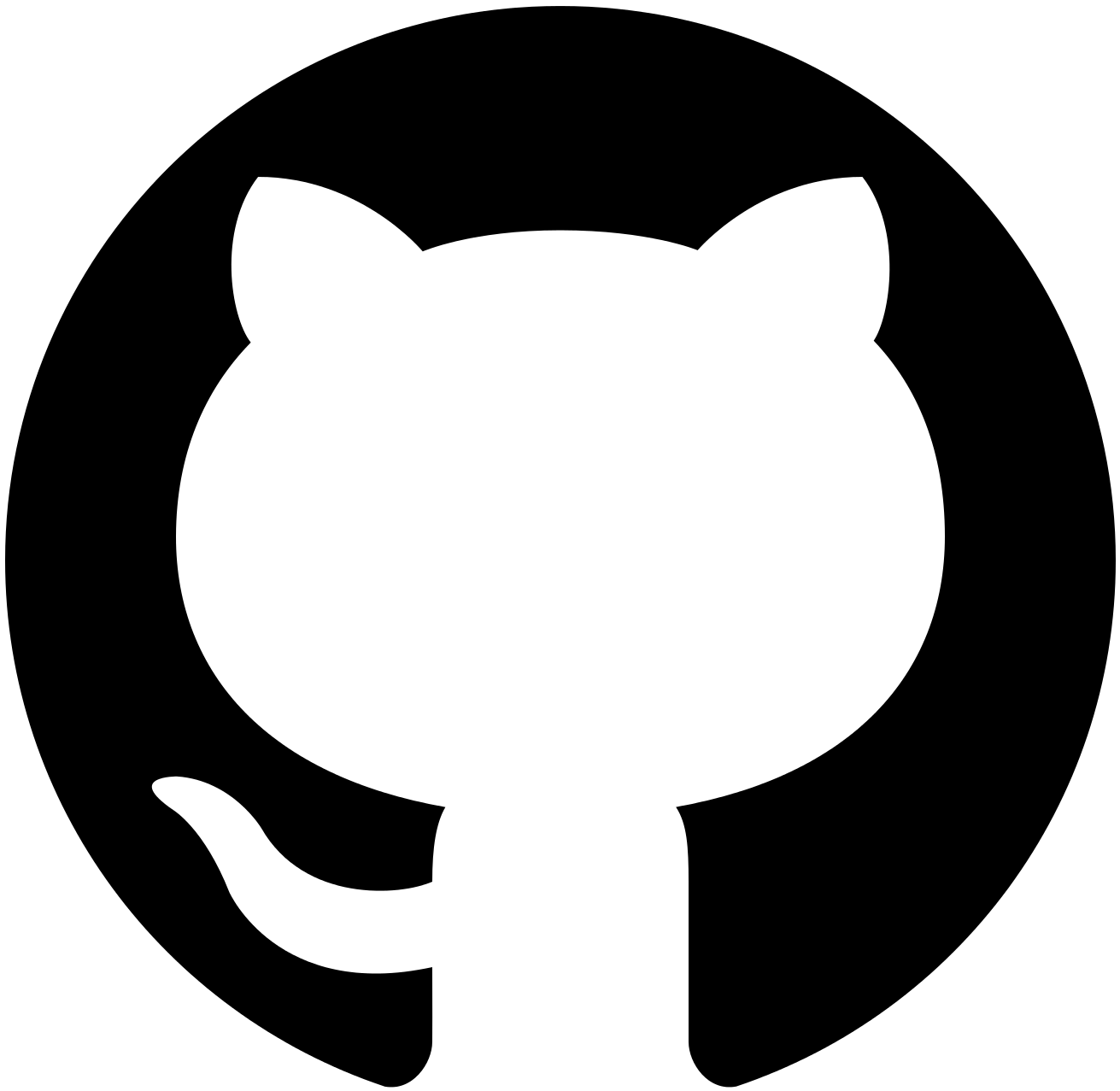}GitHub Repo]{\href{https://github.com/DAGroup-PKU/DCSAE}{\texttt{https://github.com/DAGroup-PKU/DCSAE}}}
\checkdata[\linkicon{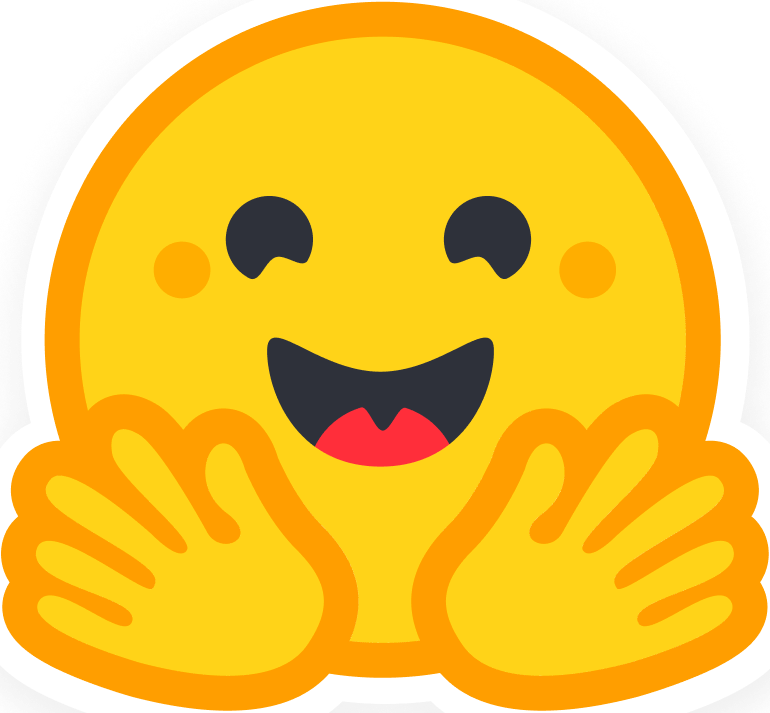}Huggingface Repo]{\href{https://huggingface.co/DAGroup-PKU/DCSAE}{\texttt{https://huggingface.co/DAGroup-PKU/DCSAE}}}
\hypersetup{
  pdftitle={DC-SAE: Deep Compression Semantic Autoencoder for Faster Diffusion Convergence},
  pdfauthor={Xu Huang, Ye Huang, Zijun Liao, Yuwei Niu, Xiaojie Li, Menghan Zhou, De Wen Soh, Xiaotong Li, Daquan Zhou}
}

\makeatletter
\fancypagestyle{firststyle}{\fancyhf{}\fancyfoot[C]{\thepage}}
\renewcommand{\@toptitlebar}{%
  \vspace*{-14mm}%
  \noindent\includegraphics[height=10mm]{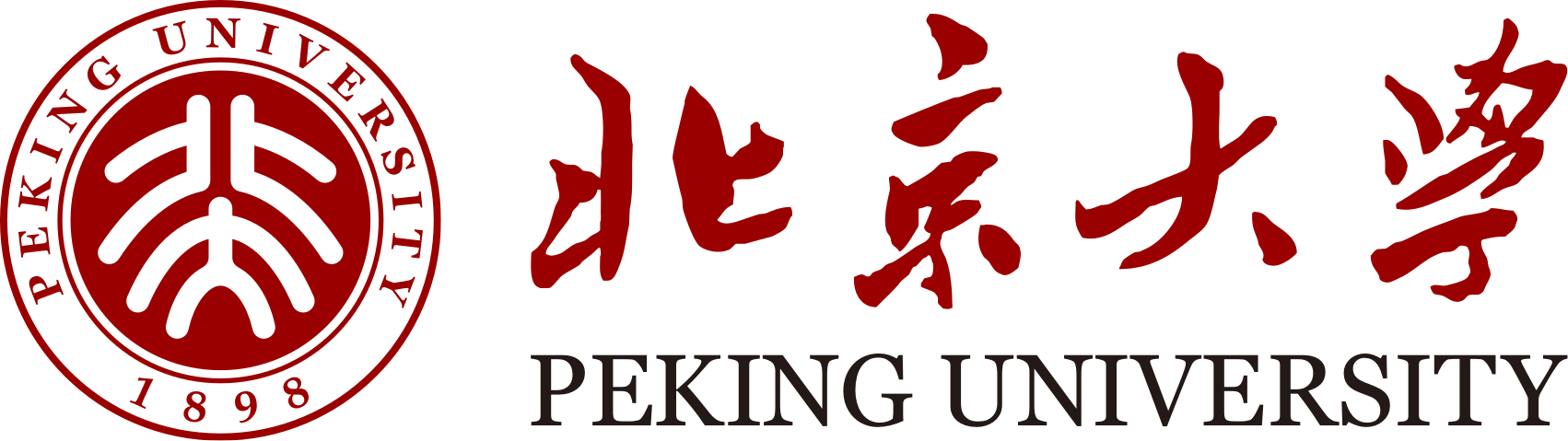}\hfill\null\par
  \vskip 1.2mm
  \pkured{\hrule height 2.5pt}
  \vskip 6mm
}
\makeatother

\begin{document}
\maketitle
\section{Introduction}

High-compression visual tokenizers such as DC-AE \cite{chen2024dcae,chen2025dcae15} are crucial because they substantially reduce the latent sequence length and computational cost of all latent-based generative models.
However, despite their high reconstruction quality, the resulting latent spaces are often less favorable for diffusion model training in terms of convergence speed: although high-compression tokenizers reduce the cost of each forward pass, diffusion models typically require more training steps to converge, which can ultimately reduce the benefits for the overall training cost.

We are therefore motivated to ask: how can we design a tokenizer that achieves both a high input compression ratio and fast diffusion model convergence?
To explore this question, we draw inspiration from recent representation autoencoders (RAEs) \cite{oquab2024dinov2, zhai2023siglip, zheng2025rae, shi2025svg}, which have been shown to enable faster diffusion model training convergence than conventional variational autoencoders (VAEs).
However, naively increasing the compression ratio of pretrained semantic encoders such as DINOv2 \cite{oquab2024dinov2} to a deep-compression regime, e.g., 32$\times$, substantially degrades the reconstruction quality of RAEs, rendering them impractical for high-fidelity generation.

\begin{figure}[t]
    \centering
    \includegraphics[width=1\linewidth]{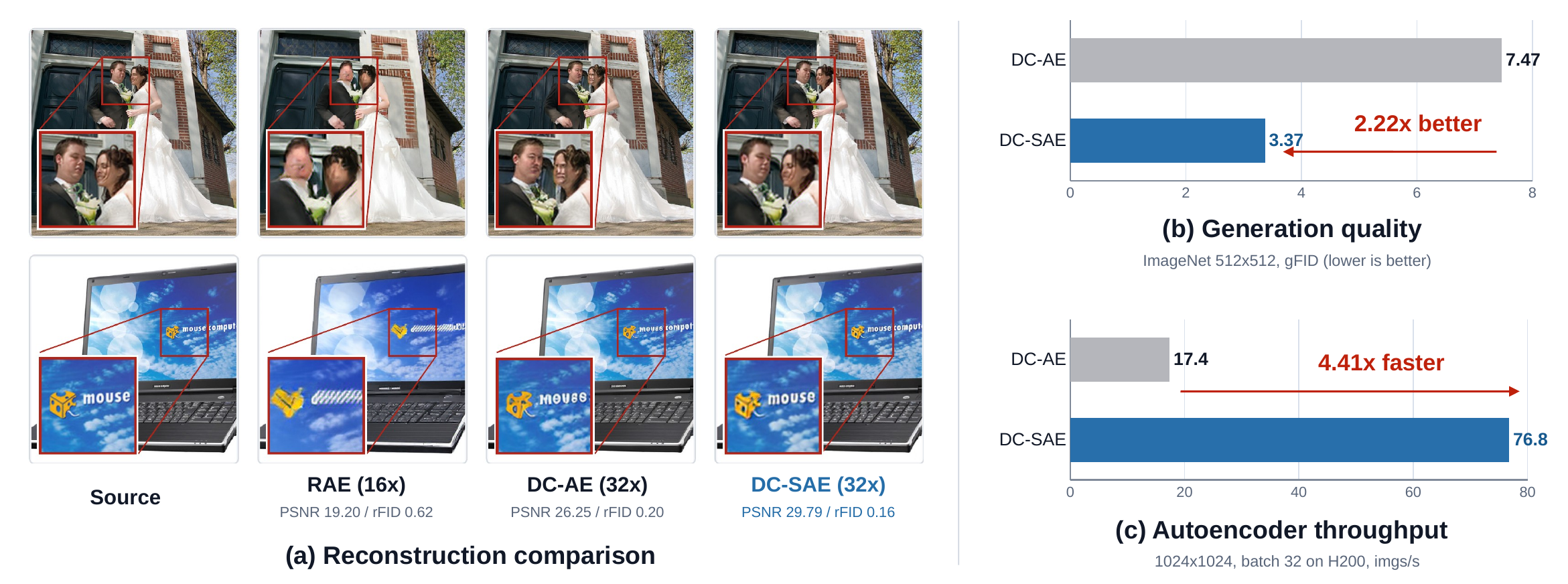}
    \caption{
    Overview of DC-SAE under $32\times$ spatial compression.
    (a) DC-SAE preserves reconstruction details missed by RAE and improves over DC-AE.
    (b) On ImageNet $512\times512$, DC-SAE achieves better gFID than DC-AE.
    (c) DC-SAE also provides substantially higher autoencoder throughput.
    }
    \label{fig:teaser}
\end{figure}

Through a detailed analysis, we find that a potential root cause of the poor reconstruction performance of RAE-like models is their lack of low-level visual information, as semantic encoders primarily capture high-level abstract representations of input images. 

We therefore propose a novel dual-stream encoder design, consisting of a semantic encoder for extracting high-level features and an unconstrained pixel-level encoder for preserving low-level visual information.
The high-level semantic features capture the major components of visual inputs and map visually similar objects to similar latent representations, yielding a more compact and structured latent space that accelerates diffusion model convergence. Meanwhile, the low-level pixel features compensate for the missing fine-grained details in semantic representations, enabling high-quality reconstruction even under a high compression ratio. The new scheme was termed \nameofmethod{}. \nameofmethod{} defines a new family of autoencoders that enjoys both a high-compression ratio and faster diffusion model convergence. 
To further improve the reconstruction quality of  \nameofmethod{}, we introduce a new module termed Spatial DeMerger. It expands the compact fused latent only on the decoder side, and thus
 keeps the generator-facing sequence short while providing denser spatial conditioning for high-fidelity reconstruction.

We conduct extensive experiments to validate the effectiveness of \nameofmethod{}. Under the challenging $32\times$ compression regime, \nameofmethod{} consistently outperforms the previous high-compression tokenizer DC-AE in both reconstruction and generation quality. Specifically, \nameofmethod{} achieves $29.79$ PSNR and $3.37$ gFID, compared with $26.25$ PSNR and $7.47$ gFID obtained by DC-AE. Beyond quality improvements, \nameofmethod{} also delivers substantially higher end-to-end throughput. As shown in Tab.~\ref{tab:throughput_comparison}, DC-SAE achieves $4.64\times$, $4.75\times$, and $4.41\times$ higher throughput than DC-AE at $256\times256$, $512\times512$, and $1024\times1024$ resolutions, respectively. These results demonstrate that \nameofmethod{} simultaneously improves reconstruction fidelity, generative performance, and computational efficiency under aggressive latent compression.
We also extend the evaluation to text-to-image generation with a $1.6$B-parameter DiT and a Qwen3-1.7B text encoder. At $1024\times1024$ resolution with classifier-free guidance (CFG), this model achieves $0.84$ on GenEval and $86.007$ on DPG-Bench, demonstrating the applicability of DC-SAE beyond class-conditional ImageNet generation.

We further provide an in-depth analysis of the proposed semantic--pixel dual-stream design. Under the $32\times$ compression setting, DiT trained on the joint semantic--pixel latent space converges substantially faster than models trained on either semantic-only or pixel-only latents. Visualization results Fig. \ref{fig:dcsae_reconstruction_source} further reveal that the pixel branch complements the semantic branch by recovering fine-grained local appearance details that are largely absent from high-level semantic representations. Together, these findings suggest that combining semantic compactness with pixel-level fidelity is key to building high-compression tokenizers that remain both reconstruction-friendly and diffusion-friendly. Our contributions are summarized as follows:
\begin{itemize}[leftmargin=*]
    \item \textbf{High-compression semantic encoders.}
    We propose a simple, yet extremely effective way to adapt the semantic autoencoders from $16\times$ to $32\times$ compression, without re-training them. This makes it possible to utilize the pre-trained semantic encoders such as series works of DINOs \cite{caron2021emerging,oquab2024dinov2,simeoni2025dinov3} and CLIPs \cite{radford2021clip,zhai2023siglip}.

    \item \textbf{Fast diffusion model training convergence.}
    We propose a novel dual-stream encoder mechanism that achieves 32$\times$ latent compression while maintaining intrinsically fast convergence for diffusion model training. Unlike prior methods that rely on explicit and often complex latent-space constraints—such as VFM alignment \cite{yao2025vavae}, self-supervised token losses \cite{li2026tcae}, or channel masking \cite{chen2025dcae15}—our approach improves convergence through the architectural design of the tokenizer itself. Moreover, it is complementary to these existing techniques and can potentially benefit from incorporating them further.
    \item \textbf{State-of-the-art deep compression autoencoder.}
    By combining these designs, we successfully train DC-SAE, which establishes a new state of the art in both reconstruction quality and diffusion model training efficiency. On ImageNet, DC-SAE improves PSNR by 13.5\% and FID by 54.9\%, while achieving a 4.41$\times$ speedup over the previous state-of-the-art deep-compression method.
\end{itemize}

\section{Related work}

\subsection{Representation encoders and unified tokenizers for generation}

Pretrained visual representation encoders have recently been adopted as alternatives to conventional VAE encoders for latent generation.
Self-supervised and vision-language models such as DINOv2, SigLIP, CLIP, and MAE provide semantically structured features that are useful for recognition and alignment~\cite{oquab2024dinov2,zhai2023siglip,radford2021clip,he2022mae}.
Representation autoencoders (RAEs) further show that replacing the reconstruction-only VAE encoder with a pretrained representation encoder can produce semantically meaningful latents and accelerate diffusion transformer training~\cite{zheng2025rae}.
However, these encoders are not originally designed for pixel-faithful reconstruction or generative tokenization.
Their objectives often discard texture, precise local geometry, color statistics, and high-frequency details that are weakly relevant for semantic understanding but crucial for reconstruction.
Complementary efforts explore semantic priors and latent-space regularization to improve latent representations and downstream generation~\cite{cui2025emu35,yao2025vavae,yue2026pae,liu2025ssvae,pan2026semantics}.

Another related line of work pursues unified tokenizers for both visual understanding and generation~\cite{qu2025tokenflow,ma2025unitok,tang2025unilip,yue2025uniflow,meituan2026longcatnext}.
TokenFlow and UniTok improve discrete token spaces through decoupled or multi-codebook designs~\cite{qu2025tokenflow,ma2025unitok}, while UniLiP and UniFlow adapt semantic features for reconstruction and generation via self-distillation or pixel-level decoding mechanisms~\cite{tang2025unilip,yue2025uniflow}.
LatentUM further studies unified multimodal modeling in a shared semantic latent space~\cite{jin2026latentum}.
Recent continuous or general-purpose visual tokenizers, such as UniFluid, Ming-UniVision, and AToken, also explore unified representations for visual understanding and generation across images, videos, or multimodal tasks~\cite{fan2025unifluid,huang2025mingunivision,lu2025atoken}.

The closest work to ours is SVG, which augments frozen DINO features with a lightweight residual branch to recover missing perceptual details for latent diffusion~\cite{shi2025svg}.
While SVG shows that VFM features can serve as a generative latent space, our focus is different: we study how to convert a frozen semantic encoder into a high-compression visual tokenizer.
Under deep compression, reconstruction becomes especially challenging because the compact semantic latent has limited spatial capacity.
DC-SAE therefore introduces an unconstrained pixel encoder to complement the frozen semantic branch with local appearance information, without relying on SVG-style self-distillation or semantic regularization.
Despite this simple design, DC-SAE improves reconstruction over SVG and recent regularized representation-tokenizer methods such as UniLiP~\cite{tang2025unilip}, while retaining the fast generation convergence of semantic latents.

\subsection{Deep-compression visual tokenizers}

Visual tokenizers compress images into compact latent representations for efficient generative modeling~\cite{rombach2022ldm}.
As image generation scales to higher resolutions and longer videos, deep-compression tokenizers have become increasingly important because the latent sequence length directly determines the cost of diffusion or transformer-based generators.
DC-AE and DC-AE 1.5 push continuous autoencoders to much higher spatial compression ratios through residual autoencoding, decoupled high-resolution adaptation, and structured latent spaces~\cite{chen2024dcae,chen2025dcae15}.
TexTok uses text descriptions to provide high-level semantic information during tokenization, allowing image tokens to focus more on fine visual details under small token budgets~\cite{zha2025textok}.
MAETok shows that masked modeling can improve latent-space structure for diffusion models without relying on variational constraints~\cite{chen2025maetok}.
TC-AE studies deep compression from the perspective of token capacity, using staged token compression and self-supervised token structuring to reduce representation collapse~\cite{li2026tcae}.
Recent image and video tokenizers further explore high-compression design from different perspectives, including efficient video autoencoding in H3AE, neural image/video tokenization in Cosmos Tokenizer, diffusion-guided decoding in DGAE, and high-fidelity discrete tokenization in WeTok~\cite{wu2025h3ae,nvidia2025cosmostokenizer,liu2025dgae,zhuang2025wetok}.
Other recent works, such as FlexTok, GloTok, and TokBench, study adaptive token allocation, global semantic structure, and tokenizer evaluation for visual generation~\cite{bachmann2025flextok,zhao2025glotok,wu2025tokbench}.

These methods reveal a common reconstruction--generation dilemma: increasing latent capacity can improve reconstruction, but may also make the latent distribution harder for downstream generators to learn~\cite{chen2025dcae15,li2026tcae}.
Our work addresses this dilemma in the setting of semantic autoencoders.
Instead of requiring a single compact latent to preserve both semantics and pixel detail, DC-SAE uses a decoupled design: a compact semantic branch provides the generator-facing representation, while an unconstrained pixel branch supplies local appearance information needed for high-fidelity reconstruction.
The two branches are concatenated and expanded by a Spatial DeMerger before decoding, which restores a denser spatial layout without requiring the generator to model a dense latent sequence.
This allows us to extend semantic autoencoders beyond the common $16\times$ regime toward higher compression, while preserving the generation efficiency benefits of compact semantic latents.

\section{Method}
\label{sec:method}

\begin{figure}[t]
    \centering
    \includegraphics[width=1\linewidth]{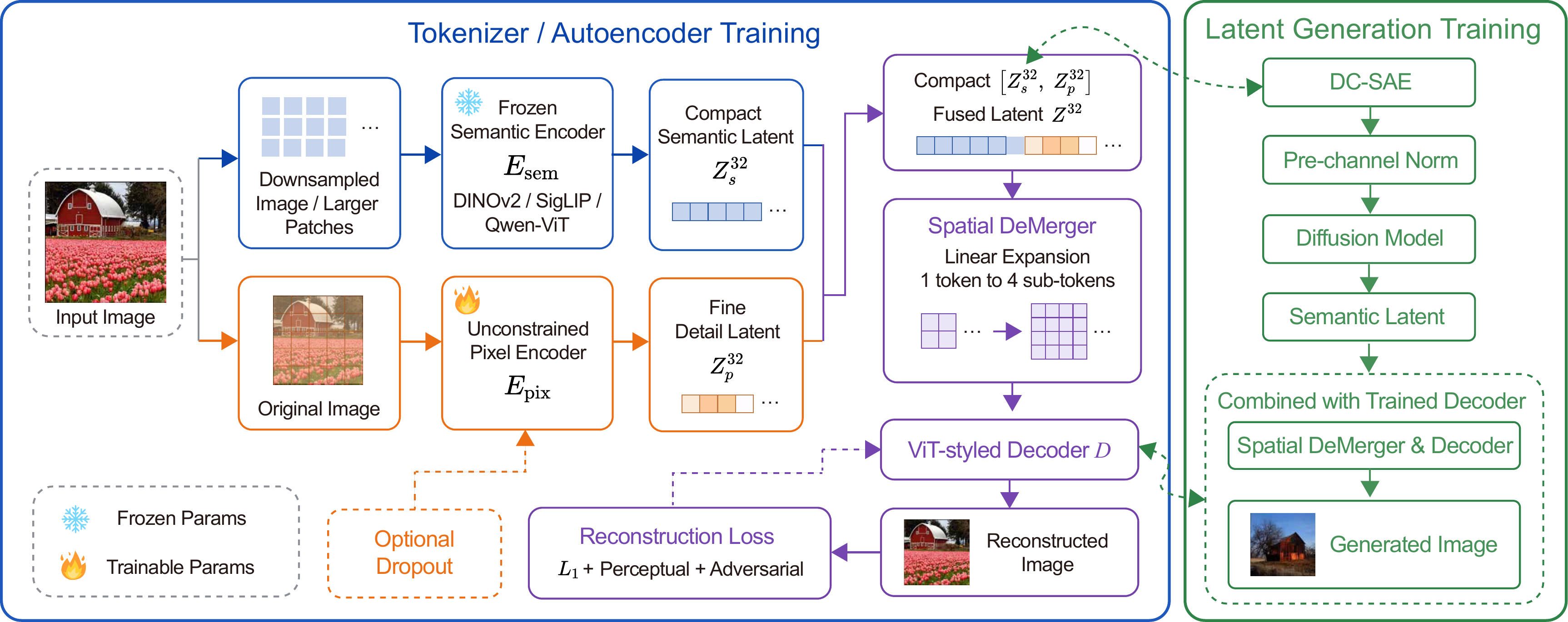}
    \caption{
    Pipeline of DC-SAE.
    We first adapt a frozen semantic encoder to a $32\times$ token grid.
    Since the resulting semantic autoencoder suffers from poor reconstruction under aggressive compression, we introduce an unconstrained pixel branch to preserve local image details.
    The semantic and pixel branches are fused and decoded by a ViT decoder with Spatial Demerger.
    }
    \label{fig:pipeline}
\end{figure}

Our goal is to build a high-compression visual tokenizer that preserves the fast convergence property of semantic representations while achieving faithful image reconstruction.
Recent semantic autoencoders show that pretrained semantic encoders provide structured latent spaces that are highly effective for diffusion training.
However, most existing semantic autoencoders operate at the standard $16\times$ spatial compression ratio.
To improve throughput, we first investigate whether such semantic autoencoders can be directly scaled to $32\times$ compression.

\subsection{Limitations of Semantic Autoencoders under Deep Compression Regime}
\label{sec:semantic_32x}

Semantic encoders are trained to preserve high-level visual structure, such as object identity, scene layout, and semantic relations.
These structures are usually stable under moderate image resizing.
Motivated by this observation, we adopt a simple strategy to construct a $32\times$ semantic autoencoder from a standard semantic encoder.
Specifically, we resize the input image to half resolution, which is equivalent to using a larger effective patch on the original image.
The resized image is then patchified, linearly projected, and fed into the frozen semantic encoder.
This directly produces a compact $32\times$ semantic latent.

The resulting semantic autoencoder is trained with the standard reconstruction objective:
\begin{equation}
    \mathcal{L}_{\mathrm{rec}}
    =
    \lambda_{1}\|\mathbf{x}-\hat{\mathbf{x}}\|_{1}
    +
    \lambda_{\mathrm{perc}}\mathcal{L}_{\mathrm{perc}}(\mathbf{x},\hat{\mathbf{x}})
    +
    \lambda_{\mathrm{adv}}\mathcal{L}_{\mathrm{adv}},
\end{equation}
where $\mathcal{L}_{\mathrm{perc}}$ denotes perceptual loss and $\mathcal{L}_{\mathrm{adv}}$ denotes the optional adversarial loss.

However, we observe that this direct strategy is insufficient for constructing an effective high-compression tokenizer.
While resizing largely preserves the global semantic structure, it substantially degrades the reconstruction fidelity.
At $256\times256$ resolution, the resulting $32\times$ semantic autoencoder achieves only around $15$ PSNR.
More importantly, the learned latent no longer exhibits the fast diffusion convergence typically expected from pretrained semantic representations.
After $80$ epochs of DiT training, it reaches only $7.15$ gFID.

One may suspect that this failure comes from the specific input-level resizing design rather than semantic compression itself.
To rule out this possibility, we further test common MLLM-style post-merge compression strategies, such as average pooling or merging already-extracted semantic tokens.
As discussed in Sec.~\ref{subsec:pre-merge_vs_post-merge}, these alternatives perform even worse than the resizing-based pre-merge strategy.
This suggests that the bottleneck is not merely the choice of compression implementation.

We attribute this failure to the objective mismatch of semantic autoencoding.
A pretrained semantic encoder is not trained for invertible reconstruction.
It keeps information useful for recognition and representation learning, but discards many pixel-level details, such as color statistics, textures, local boundaries, and high-frequency patterns.
When the compression ratio is increased from $16\times$ to $32\times$, this information loss is further amplified.
Therefore, a semantic branch alone is insufficient for high-compression visual tokenization.

\begin{figure}[t]
\centering
\begin{minipage}[t]{\linewidth}
\centering
\includegraphics[width=\linewidth]{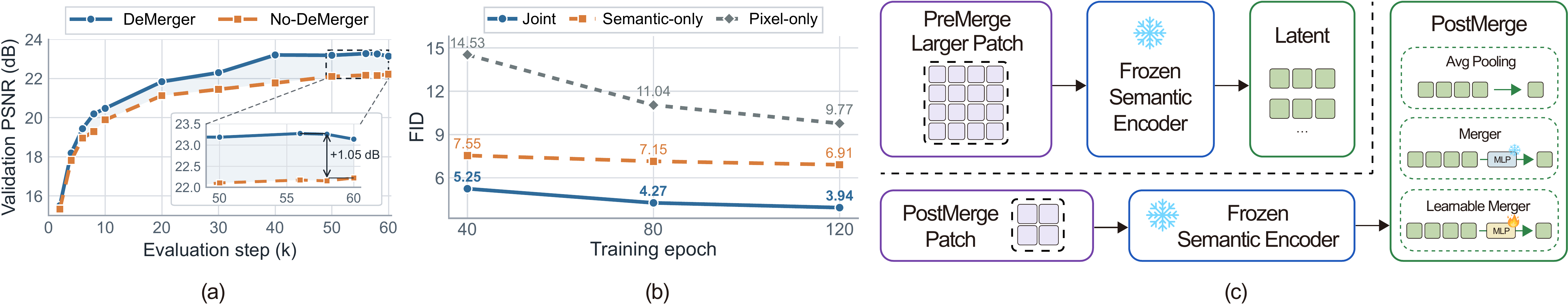}
\caption{
(a) Validation PSNR comparison between DeMerger and No-DeMerger variants. 
(b) DiT training convergence on joint semantic--pixel, semantic-only, and pixel-only latents under $32\times$ compression. 
(c) Illustration of pre-merge and post-merge semantic compression for constructing a $32\times$ semantic tokenizer. 
}
\label{fig:merge_figure}
\end{minipage}
\end{figure}

\subsection{Semantic-guided Unconstrained AutoEncoder}
\label{sec:pixel_branch}

The failure of the direct $32\times$ semantic autoencoder shows that semantic latents alone are insufficient for high-fidelity reconstruction under aggressive compression.
A straightforward way to improve reconstruction is to use an unconstrained, high-channel autoencoder, since a larger latent capacity can preserve more pixel-level information.
However, prior works have shown that such reconstruction-oriented latents are often difficult for diffusion models to learn.
To improve latent generability, existing methods usually introduce additional constraints or regularization, such as KL regularization, representation alignment, structured latent design, or channel masking~\cite{kingma2022autoencodingvariationalbayes,rombach2022ldm,yao2025vavae,chen2025dcae15}.
This reflects a common trade-off: increasing autoencoder capacity improves reconstruction, but may hurt diffusion convergence.

Our key motivation is that this trade-off can be alleviated by semantic guidance.
Recent works suggest that semantic representation latents can provide useful structure for diffusion training, and Latent Forcing further indicates that semantic latents can guide pixel-level generation trajectories~\cite{baade2026latentforcingreorderingdiffusion}.
Therefore, we hypothesize that, even under high compression, a frozen semantic encoder can serve as a structural guide for an unconstrained high-channel pixel autoencoder.
Instead of constraining the pixel latent itself, we pair it with a semantic latent and jointly train the two branches inside a single autoencoder.

As shown in Fig.~\ref{fig:pipeline}, DC-SAE introduces an unconstrained pixel encoder in parallel with the frozen $32\times$ semantic branch.
The semantic branch provides a compact and structured latent space that captures object-level layout and semantic relations.
The pixel branch is trained from scratch and focuses on reconstruction-critical details, such as texture, color statistics, local boundaries, and high-frequency information.
The two latents are concatenated along the channel dimension, fused by a lightweight projection, and decoded jointly.
In this way, the final latent space combines the generative structure of semantic representations with the reconstruction capacity of unconstrained pixel features.
Since the pixel branch is directly optimized for reconstruction, the decoder may over-rely on it and under-utilize the frozen semantic branch.
We therefore apply sample-wise dropout to the pixel latent during tokenizer training, forcing the decoder to use semantic structure when the pixel branch is dropped and to treat the pixel branch as complementary detail when it is present.

This decomposition leads to a more suitable latent space for high-compression diffusion training.
As shown in Fig.~\ref{fig:merge_figure}(b), the joint semantic--pixel latent converges faster than both semantic-only and pixel-only alternatives, reaching $4.27$ gFID after $80$ epochs compared with $7.15$ and $11.04$, respectively.
This indicates that neither branch alone is sufficient: semantic features provide a structured, diffusion-friendly representation but lose reconstruction-critical details, whereas pixel latents preserve such details but are harder to model at high channel capacity.
By combining them, DC-SAE obtains a latent representation that supports both faithful reconstruction and efficient generative modeling.

After tokenizer training, the entire tokenizer is frozen, and DiT is trained on the fused latent.
Because the semantic and pixel branches are jointly optimized through the same decoder, the resulting latent is more generation-friendly than simply concatenating an independently trained pixel autoencoder with semantic features.

\subsection{ViT Decoder with Spatial Demerger}
\label{sec:spatial_demerger}

The semantic--pixel latent introduced above improves the information content of the tokenizer under $32\times$ compression.
However, the compact latent grid also creates a decoder-side spatial bottleneck.
For diffusion modeling, a sparse $32\times$ grid is desirable because it significantly reduces the sequence length.
For reconstruction, however, this grid is less suitable for a ViT decoder, which relies on spatial tokens and attention to recover local image details.
A denser decoder-side token grid can thus provide more spatial positions for modeling textures, boundaries, and high-frequency structures.

To address this mismatch, we introduce Spatial DeMerger before the ViT decoder.
Spatial DeMerger expands each compact latent token into a small group of spatial sub-tokens before decoding.
Importantly, this expansion is applied only on the decoder side.
The DiT still operates on the original compact $32\times$ compressed latent, so the efficiency benefit of high compression is preserved, while the decoder receives a denser grid for high-fidelity reconstruction.

This design decouples the latent grid used for diffusion modeling from the token grid used for image decoding.
The compact grid keeps DiT training and sampling efficient, whereas the expanded decoder-side grid improves the reconstruction ability of the ViT decoder.
In Sec.~\ref{sec:ablation}, we further provide ablation studies showing that Spatial DeMerger consistently improves reconstruction quality under high compression.

\begin{table}[t]
\centering
\caption{
Comparison of $16\times$ semantic autoencoders on reconstruction quality and ImageNet $256 \times 256$ generation after 80 epochs.
DC-SAE-16x achieves strong reconstruction while retaining competitive generation efficiency, making it a flexible baseline for high-compression tokenizer design.
\emph{Notice:} The $16\times$ DC-SAE setting is used only in this ablation; all other DC-SAE results in this paper use the $32\times$ setting.
}
\label{tab:sae_16x}
\small
\setlength{\tabcolsep}{6pt}
\renewcommand{\arraystretch}{1.12}
\resizebox{0.85\linewidth}{!}{
\begin{tabular}{lcccccc}
\toprule
Metric
& RAE~\cite{zheng2025rae}
& SD-VAE~\cite{rombach2022ldm}
& VA-VAE~\cite{yao2025vavae}
& SVG~\cite{shi2025svg}
& REPA~\cite{yu2024repa}
& \textbf{DC-SAE-16x} \\
\midrule
\rowcolor{blue!8}\multicolumn{7}{l}{\emph{Reconstruction}} \\
PSNR $\uparrow$
& 19.20
& 24.08
& \textbf{27.96}
& 23.89
& --
& \underline{27.80} \\
rFID $\downarrow$
& 0.62
& 0.87
& \underline{0.28}
& 0.65
& --
& \textbf{0.26} \\
\midrule
\rowcolor{blue!8}\multicolumn{7}{l}{\emph{Generation}} \\
gFID $\downarrow$
& \textbf{2.16}
& --
& 5.96
& 6.57
& 7.90
& \underline{3.09} \\
IS $\uparrow$
& \textbf{214.80}
& --
& 128.00
& 137.90
& 118.60
& \underline{189.95} \\
\bottomrule
\end{tabular}
}
\end{table}

\begin{figure}[t]
    \centering
    \includegraphics[width=1\linewidth]{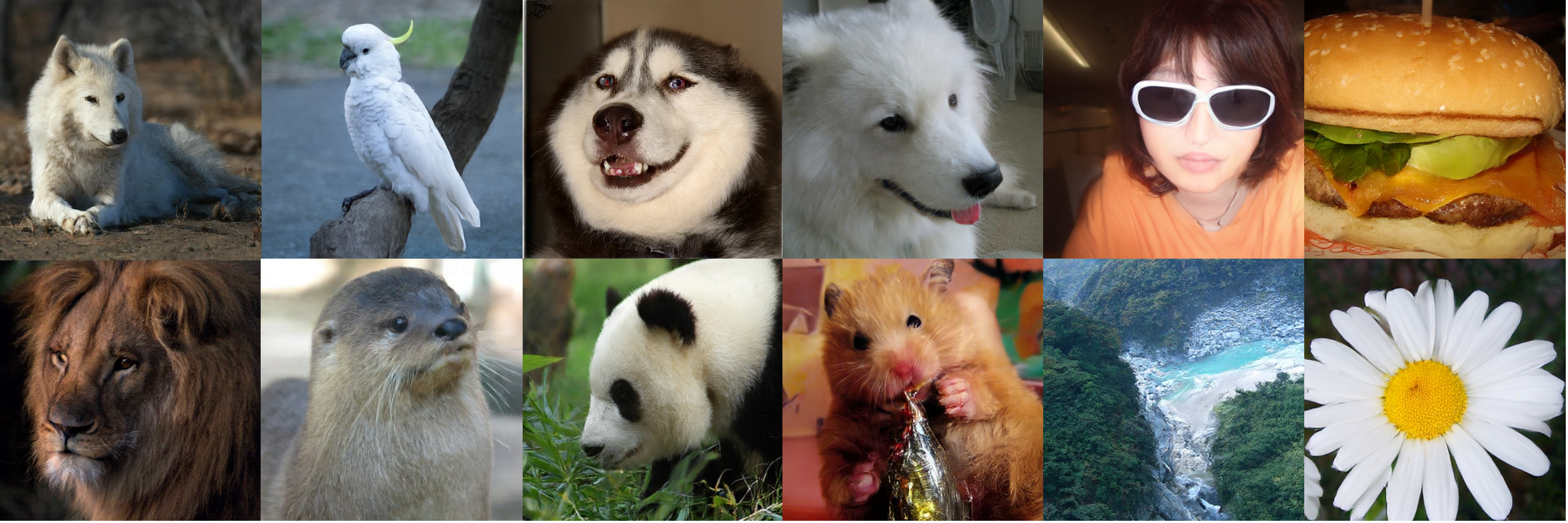}
    \caption{
    Qualitative results on ImageNet. The images are generated with  $512\times512$ resolutions in the latent space of DC-SAE under $32\times$ compression, with randomly sampled class conditions.
    }
    \label{fig:generation_sample}
\end{figure}

\Needspace{6\baselineskip}
\section{Experiments}
\label{sec:exp}

\begin{table}[t]
\centering
\caption{
Comparison with state-of-the-art image generative models on ImageNet $512\times512$ class-conditional generation.
DC-SAE uses a $32\times$ high-compression semantic tokenizer and trains a DiT-XL generator in the learned latent space.
}
\label{tab:imagenet512_sota}
\small
\setlength{\tabcolsep}{4pt}
\resizebox{\linewidth}{!}{%
\begin{tabular}{llcccccccc}
\toprule
\multicolumn{9}{l}{\textbf{ImageNet $512\times512$}} \\
\midrule
Image Generative Model 
& Autoencoder 
& Compression Rate 
& Params (B) 
& \multicolumn{2}{c}{gFID $\downarrow$} 
& Inception Score $\uparrow$ 
& PSNR $\uparrow$ 
& rFiD $\downarrow$ \\
\cmidrule(lr){5-6}
& & & & w/o CFG & w/ CFG & & \\
\midrule
DiT-XL~\cite{peebles2023scalable} 
& Flux-VAE-f8c16~\cite{flux2024} 
& $8\times$ 
& 0.68 
& 27.35 
& 8.72 
& 53.09 
& --
& -- \\

\midrule
DiT-XL~\cite{peebles2023scalable} 
& SD-VAE-f8c4~\cite{rombach2022ldm} 
& $8\times$ 
& 0.67 
& 12.03 
& 3.04 
& 105.25 
& \multirow{2}{*}{--}
& \multirow{2}{*}{--} \\

SiT-XL~\cite{ma2024sit} 
& SD-VAE-f8c4~\cite{rombach2022ldm} 
& $8\times$ 
& 0.67 
& -- 
& 2.62 
& -- 
&
& \\

\midrule
MAGVIT-v2~\cite{yu2023language} 
& -- 
& -- 
& -- 
& 3.07 
& 1.91 
& 213.10 
& --
& --\\

MAR-L~\cite{li2024autoregressiveimagegenerationvector} 
& -- 
& -- 
& -- 
& 2.74 
& 1.73 
& 205.20 
& --
& -- \\

EDM2-XXL~\cite{Karras2024edm2} 
& -- 
& -- 
& -- 
& 1.91 
& 1.81 
& -- 
& --
& -- \\

\midrule
SiT-XL~\cite{ma2024sit} 
& DC-AE-f32c32~\cite{chen2024dcae} 
& $32\times$ 
& 0.67 
& 7.47 
& 2.41 
& 131.37 
& \multirow{3}{*}{26.25}
& \multirow{3}{*}{0.20} \\

USiT-H~\cite{chen2024dcae} 
& DC-AE-f32c32~\cite{chen2024dcae} 
& $32\times$ 
& 0.50 
& 3.80 
& 1.89 
& 174.58 
&
& \\

USiT-2B~\cite{chen2024dcae} 
& DC-AE-f32c32~\cite{chen2024dcae} 
& $32\times$ 
& 1.58 
& 2.90 
& 1.72 
& 187.68 
&
& \\

\midrule
\rowcolor{blue!8}
\textbf{DiT-XL {\scriptsize\mdseries + DDT wide head \cite{wang2025ddtdecoupleddiffusiontransformer}}}
& \textbf{DC-SAE} 
& $\mathbf{32\times}$ 
& 0.83
& \textbf{3.37}
& 1.99
& \textbf{205.83} 
& \textbf{29.79} & \textbf{0.16} \\

\bottomrule
\end{tabular}
\vphantom{\rule{0pt}{1pt}}%
}
\end{table}

\subsection{Experimental Setting}

We evaluate DC-SAE on class-conditional ImageNet generation at both $256\times256$ and $512\times512$ resolutions, as well as text-to-image generation at $1024\times1024$ resolution. 
For ImageNet tokenizer training, we first train DC-SAE at $256\times256$ resolution and then finetune it at $512\times512$ resolution under the target spatial compression ratio. 
Detailed tokenizer training settings are provided in Appendix~\ref{app:tokenizer_details}. 
After tokenizer training, the tokenizer is frozen, and a latent diffusion/transformer generator is trained in the learned latent space. 
For ImageNet diffusion training, we follow the training protocol of DC-AE~\cite{chen2024dcae}. 
We report reconstruction quality using rFID, and ImageNet generation quality using gFID and Inception Score. Text-to-image training and evaluation are detailed in Sec.~\ref{sec:t2i}.

\subsection{Main Result}
\paragraph{ImageNet $256\times 256$ generation}

We first compare DC-SAE with DC-AE and DC-AE-1.5 under the ImageNet $256\times256$ setting. 
This setting evaluates whether the proposed semantic--pixel decoupled tokenizer can improve generation quality under the same high-compression regime. 
As shown in Tab.~\ref{tab:imagenet256_comparison}, DC-SAE improves the reconstruction rFID from $0.69$ to $0.45$ and achieves a gFID of $3.31$ with an Inception Score of $180.71$ using a DiT-XL generator. 
Compared with DC-AE and DC-AE-1.5 under the same DiT-XL backbone, DC-SAE provides substantially better generation quality, showing that the proposed tokenizer produces a more generation-friendly latent space while preserving reconstruction fidelity.
These results suggest that DC-SAE is already a strong high-compression tokenizer at $256\times256$ resolution. 
Although our generator uses the standard DiT-XL backbone rather than the specialized USiT architecture, it achieves generation quality comparable to or better than stronger DC-AE baselines, while also improving reconstruction rFID. Following RAE, we also use a DiT-XL variant with a DDT-wide head for diffusion training~\citep{wang2025ddtdecoupleddiffusiontransformer}.

\paragraph{ImageNet $512\times512$ generation}

We further evaluate DC-SAE on ImageNet $512\times512$ class-conditional generation. 
The tokenizer is finetuned at $512\times512$ resolution under a $32\times$ spatial compression ratio, and a DiT-XL generator is trained in the frozen DC-SAE latent space for 400 epochs following the DC-AE training setting. 
As shown in Tab.~\ref{tab:imagenet512_sota}, DC-SAE achieves a gFID of $3.37$ without CFG under the $512\times512$ setting, demonstrating that the proposed tokenizer scales to higher-resolution generation while maintaining a compact latent representation. We provide qualitative generation samples in Fig.~\ref{fig:generation_sample}.

\subsection{Text-to-Image Generation}
\label{sec:t2i}

\paragraph{Experimental setup.}
To evaluate DC-SAE beyond class-conditional ImageNet generation, we train a text-to-image model with a $1.6$B-parameter DiT backbone and Qwen3-1.7B~\cite{qwen3technicalreport} as the text encoder, using DC-SAE at $32\times$ spatial compression.
Training uses a subset of internal data together with LAION-COCO~\cite{schuhmann2022laioncoco}.
We progressively pretrain the model at $256\times256$ resolution for $100$K iterations with a batch size of $1024$, followed by another $100$K iterations at $512\times512$ with the same batch size.
We then perform supervised fine-tuning (SFT) at $1024\times1024$ resolution for $40$K iterations with a batch size of $256$.
The final model is evaluated with classifier-free guidance (CFG) on GenEval~\cite{ghosh2023geneval} and DPG-Bench~\cite{hu2024ella}, which assess compositional text-to-image alignment and dense-prompt following, respectively.

\begin{table}[tbp]
\centering
\caption{Text-to-image generation results with CFG. The DC-SAE model is evaluated at $1024\times1024$ resolution. Generator sizes and tokenizer compression ratios are listed explicitly; the baselines use different model and training configurations. Higher scores are better for both benchmarks.}
\label{tab:t2i_results}
\small
\setlength{\tabcolsep}{4pt}
\resizebox{\linewidth}{!}{%
\begin{tabular}{@{}llcccc@{}}
\toprule
Image Generative Model & Autoencoder & Compression Rate & Params (B) & GenEval $\uparrow$ & DPG-Bench $\uparrow$ \\
\midrule
DC-Gen-FLUX.1-Krea-12B~\cite{he2026dc} & DC-AE f32c32~\cite{chen2024dcae} & $32\times$ & 12 & 0.72 & \textbf{87.073} \\
DC-Gen-FLUX.1-Krea-12B~\cite{he2026dc} & DC-AE-1.5 f64c128~\cite{chen2025dcae15} & $64\times$ & 12 & 0.59 & 75.439 \\
\midrule
\rowcolor{blue!8}
\textbf{1.6B-DiT} & \textbf{DC-SAE} & $32\times$ & 1.6 & \textbf{0.84} & 86.007 \\
\bottomrule
\end{tabular}
}
\end{table}

\paragraph{Results.}
As shown in Tab.~\ref{tab:t2i_results}, our model achieves $0.84$ on GenEval, exceeding the DC-AE and DC-AE-1.5-based DC-Gen-FLUX.1-Krea-12B baselines by $0.12$ and $0.25$, respectively.
On DPG-Bench, our model obtains $86.007$, compared with $87.073$ for the DC-AE baseline and $75.439$ for the DC-AE-1.5 baseline.
Thus, the $1.6$B-parameter generator achieves the highest GenEval score among these models and remains competitive on DPG-Bench against the $12$B DC-AE-based generator.
These results support the applicability of DC-SAE to open-ended text-to-image generation with a compact latent grid.

\subsection{Ablation Studies}
\label{sec:ablation}

We conduct several ablation studies to understand the four design choices of DC-SAE.
Specifically, we ask:
whether the proposed semantic--pixel decoupled design is effective under the standard $16\times$ setting,
how to adapt pretrained semantic encoders to the $32\times$ token grid,
whether Spatial DeMerger is necessary for high-compression ViT decoding,
and whether the gain comes from joint semantic--pixel autoencoding rather than simply exposing DiT to semantic features.

\paragraph{Validation under $16\times$ compression.}
We first evaluate the proposed semantic--pixel decoupled design under the standard $16\times$ setting.
This setting isolates the effect of the pixel branch, since pretrained semantic encoders already provide a sufficiently dense token grid at $16\times$ compression.
As shown in Tab.~\ref{tab:sae_16x}, DC-SAE substantially improves reconstruction quality over semantic-only autoencoders.
Compared with RAE, which obtains around $19$ PSNR, DC-SAE reaches $27.80$ PSNR and $0.26$ rFID.
Meanwhile, its latent space remains effective for diffusion training, achieving strong ImageNet $256\times256$ generation performance after $80$ epochs. This result confirms the basic motivation of our design.
The frozen semantic branch provides a structured latent space that benefits generation, but it lacks reconstruction-critical details.
Adding a trainable pixel branch closes most of this reconstruction gap without destroying the diffusion-friendly structure of the semantic latent.

\begin{table}[tbp]
\centering
\caption{
Comparison between post-merge and pre-merge strategies for constructing a $32\times$ semantic tokenizer.
}
\label{tab:post_merge_32x}
\scriptsize
\renewcommand{\bffont}{\scriptsize}
\setlength{\tabcolsep}{3pt}
\resizebox{0.8\linewidth}{!}{%
\begin{tabular}{lcccc}
\toprule
Semantic Encoder & PSNR $\uparrow$ & rFID $\downarrow$ & gFID $\downarrow$ & IS $\uparrow$ \\
\midrule
\multicolumn{5}{l}{\textbf{Post-merge}} \\
Qwen-ViT Frozen Merger & 23.40 & 0.62 & 25.96 & 66.83 \\
Qwen-ViT AvgPool Merger & 23.07 & 0.61 & 17.71 & 84.45 \\
Qwen-ViT Learnable Merger & 23.85 & 0.58 & 16.84 & 78.34 \\
DINOv2 Learnable Merger & 24.03 & 0.35 & 8.60 & 108.11 \\
\midrule
\multicolumn{5}{l}{\textbf{Pre-merge}} \\
DINOv2 & 23.13 & 0.45 & \textbf{3.31} & \textbf{180.71} \\
\bottomrule
\end{tabular}
}
\end{table}

\begin{table}[tbp]
\centering
\caption{
Comparison with DC-AE and DC-AE-1.5 on ImageNet $256\times256$ class-conditional generation.
}
\label{tab:imagenet256_comparison}
\setlength{\tabcolsep}{3pt}
\resizebox{\linewidth}{!}{%
\begin{tabular}{llcccc}
\toprule
\multicolumn{6}{l}{\textbf{ImageNet $256\times256$}} \\
\midrule
Diffusion Model 
& Setting 
& Autoencoder 
& rFID $\downarrow$ 
& gFID $\downarrow$ 
& Inception Score $\uparrow$ \\
\midrule
DiT-XL~\cite{peebles2023scalable} 
& f32c32 
& DC-AE~\cite{chen2024dcae} 
& 0.69 
& 10.18 
& 107.49 \\

DiT-XL~\cite{peebles2023scalable} 
& f32c128
& DC-AE~\cite{chen2024dcae} 
& 0.26 
& 26.44 
& 53.41 \\
DiT-XL~\cite{peebles2023scalable} 
& f32c32 
& DC-AE-1.5~\cite{chen2025dcae15} 
& -- 
& 10.50 
& 107.99 \\

DiT-XL~\cite{peebles2023scalable} 
& f32c128
& DC-AE-1.5~\cite{chen2025dcae15} 
& 0.26 
& 17.31 
& 80.38 \\

\textbf{DiT-XL \cite{peebles2023scalable} } 
& \textbf{f32c64} 
& \textbf{DC-SAE (Ours)} 
& \textbf{0.45} 
& \textbf{5.67} 
& \textbf{156.68} \\
\midrule
USiT-H~\cite{chen2024dcae} 
& f32c32 
& DC-AE~\cite{chen2024dcae} 
& 0.69 
& 3.89 
& 167.40 \\
USiT-H~\cite{chen2025dcae15} 
& f32c32 
& DC-AE-1.5~\cite{chen2025dcae15} 
& -- 
& 4.15 
& 166.94 \\

\midrule
DiT-XL ~\cite{peebles2023scalable} 
& f32c128 
& TC-AE ~\cite{li2026tcae}
& 0.35 
& 7.16
& - \\
\midrule
\textbf{DiT-XL {\scriptsize\mdseries + DDT wide head \cite{wang2025ddtdecoupleddiffusiontransformer}}}& \textbf{f32c64} 
& \textbf{DC-SAE (Ours)} 
& \textbf{0.45} 
& \textbf{3.31} 
& \textbf{180.71} \\
\bottomrule
\end{tabular}
}
\end{table}

\paragraph{Pre-merge versus post-merge semantic compression.}
\label{subsec:pre-merge_vs_post-merge}
We next compare different ways of adapting pretrained semantic encoders to $32\times$ compression.
Motivated by recent MLLM efforts on high-compression vision tokenizers, a straightforward approach is to first extract dense visual tokens and then reduce the token count through local token merging. As shown in Fig.~\ref{fig:merge_figure}(c).
In our setting, this corresponds to first extracting dense $16\times$ semantic tokens and then merging each local token group into a compact $32\times$ token.
We refer to this family of methods as \emph{post-merge compression}, since compression is applied after semantic feature extraction.
We evaluate several variants, including average pooling \cite{guo2025seed1,gemma4}, frozen merger modules \cite{bai2025qwen3vltechnicalreport}, and learnable mergers.
Although these methods are simple and commonly used for reducing visual token count, they are not ideal for generative tokenization.
As shown in Tab.~\ref{tab:post_merge_32x}, post-merge variants consistently underperform.
For example, Qwen-ViT, with its frozen merger, obtains poor generation quality, and even learnable merger variants remain limited.
This indicates that reducing tokens after semantic abstraction discards information that is difficult to recover later.
In contrast, the resizing-based strategy used in Sec.~\ref{sec:semantic_32x} compresses the input before semantic abstraction.
We refer to this strategy as \emph{pre-merge compression}, since the semantic encoder directly forms representations under the target $32\times$ token budget rather than compressing already-extracted dense semantic tokens.
Empirically, this strategy leads to better generation quality, showing that compression should be introduced before semantic encoding.
\paragraph{Effectiveness of Spatial DeMerger.}
Finally, we evaluate the role of Spatial DeMerger in the decoder.
At $32\times$ compression, the latent grid is highly compact, which is favorable for diffusion throughput but challenging for reconstruction.
This issue is particularly important for ViT decoders, since they benefit from a denser spatial token grid for local detail recovery.

As shown in Fig.~\ref{fig:merge_figure}(a), under the standard DINOv2 pre-merge setting on ImageNet $256\times256$ with $32\times$ compression, adding Spatial DeMerger consistently improves validation PSNR over the No-DeMerger variant.
This supports our motivation that the compact diffusion latent and the decoder-side token grid should be treated differently: the diffusion model should operate on the sparse $32\times$ latent for efficiency, while the decoder benefits from a denser grid for high-fidelity reconstruction.
Additional results with QwenViT-based semantic encoders are provided in Appendix~\ref{sec:appendix_spatial_demerger_analysis}.

\paragraph{Joint semantic--pixel training versus late concatenation.}

A natural question is whether the gain of DC-SAE comes from joint semantic--pixel autoencoding, or simply from exposing DiT to semantic features.
To isolate this factor, we compare two settings in Tab.~\ref{tab:standalone_ae_ablation}: \textit{Joint} and \textit{Late-Concat}.
\textit{Joint} is the DC-SAE baseline, where the decoder reconstructs images from the concatenated semantic and pixel latents.
Thus, the pixel branch is learned together with the semantic branch inside the autoencoder.
In contrast, \textit{Late-Concat} first trains a pixel autoencoder alone, whose decoder only receives the pixel latent.
The frozen DINOv2 semantic latent is concatenated with the pixel latent only later, during DiT training.
Therefore, DiT receives semantic features in both settings, but only \textit{Joint} learns a coupled semantic--pixel latent space during autoencoding.

\begin{wraptable}{r}{0.48\textwidth}
\centering
\scriptsize
\caption{
Standalone autoencoder ablation.
Entries are FID / IS after DiT training.
}
\label{tab:standalone_ae_ablation}
\setlength{\tabcolsep}{3pt}
\resizebox{\linewidth}{!}{%
\begin{tabular}{lccc}
\toprule
Metric & Joint & Late-Concat & Late-Concat-E \\
\midrule
PSNR $\uparrow$ 
& 29.79 
& 31.91 
& 29.15 \\
rFID $\downarrow$ 
& 0.17 
& 0.09 
& 0.17 \\
\midrule
20 ep. 
& 8.12 / 134.23 
& 10.95 / 122.48 
& 10.35 / 127.78 \\
40 ep. 
& 5.23 / 166.76 
& 7.41 / 153.16 
& 6.91 / 156.97 \\
60 ep. 
& 4.33 / 180.31 
& 6.22 / 168.13 
& 6.01 / 167.48 \\
80 ep. 
& 4.02 / 185.46 
& 5.64 / 174.52 
& 5.40 / 173.00 \\
\bottomrule
\end{tabular}
}
\end{wraptable}

As shown in Tab.~\ref{tab:standalone_ae_ablation}, late concatenation does not match the joint training variant.
Although \textit{Late-Concat} achieves better reconstruction quality than \textit{Joint}, it gives worse generation performance, with an 80-epoch FID of $5.64$ versus $4.02$.
To rule out a reconstruction--generation conflict, \textit{Late-Concat-E} uses an earlier checkpoint of the same pixel autoencoder, whose reconstruction quality is closer to \textit{Joint}.
It still underperforms, reaching an 80-epoch FID of $5.40$.
These results indicate that DC-SAE's gain is not merely from adding semantic features to DiT, but from jointly training the decoder to use semantic and pixel latents as a compatible representation for generation.

\newpage
\subsection{Discussion: Understanding Component Contributions to Reconstruction Fidelity}
\label{subsec:reconstruction_source}
\noindent
\begin{minipage}[t]{0.49\linewidth}
\vspace{0pt}
\setlength{\parskip}{6pt}
We further analyze where the reconstruction ability of DC-SAE comes from.
Since DC-SAE contains both a semantic branch and a pixel branch, the decoder may rely on either branch for reconstruction.
To study this, we reconstruct images using the full latent, the latent with the semantic component masked, and the latent with the pixel component masked.

As shown in Figure~\ref{fig:dcsae_reconstruction_source}, masking out the semantic component has little effect on reconstruction quality, while masking out the pixel component causes severe degradation.
This suggests that high-fidelity reconstruction mainly comes from the pixel branch.
The semantic branch, in contrast, primarily provides high-level structure and a generation-friendly representation space rather than detailed appearance information.

This finding is consistent with the design of DC-SAE.
The pixel branch preserves local texture, color, and spatial details for reconstruction, while the semantic branch guides the joint latent toward a structure that is easier for DiT to learn.
Thus, DC-SAE does not require the frozen semantic encoder to carry all low-level information; instead, reconstruction and generative structure are handled by two complementary branches.
\end{minipage}
\hfill
\begin{minipage}[t]{0.48\linewidth}
\vspace{0pt}
\captionsetup{hypcap=false}
    \centering
    \includegraphics[width=\linewidth]{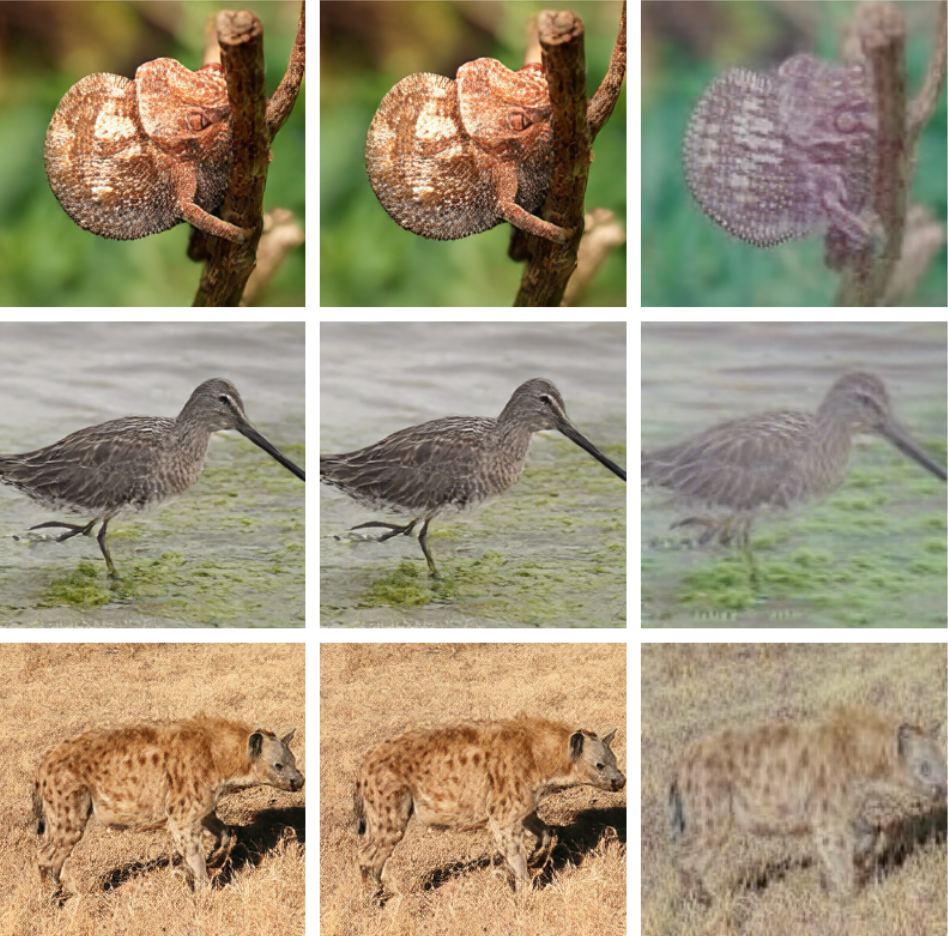}
    \captionof{figure}{
    Investigating where DC-SAE's reconstruction ability comes from.
    We compare three reconstruction settings: using the full latent, masking the semantic component, and masking the pixel component.
    Masking the semantic branch has little impact on reconstruction quality, whereas masking the pixel branch causes severe degradation.
    This indicates that DC-SAE primarily relies on the pixel branch for reconstruction.
    }
    \label{fig:dcsae_reconstruction_source}
\end{minipage}
\par\medskip

\section{Conclusion}

We propose DC-SAE, a decoupled compact semantic autoencoder for high-compression latent image generation. 
DC-SAE addresses the reconstruction--generation tradeoff by combining a frozen semantic encoder for generation-friendly structure with an unconstrained pixel branch for low-level reconstruction details. 
We further identify post-hoc semantic token merging as a key obstacle to extending semantic autoencoders to $32\times$ compression, as it discards spatial and low-level details after semantic abstraction. 
We address this with input-level semantic compression and a decoder-side Spatial DeMerger, enabling high-compression semantic tokenization with strong reconstruction fidelity.
Experiments on ImageNet show that DC-SAE achieves strong reconstruction quality, faster diffusion convergence, and competitive generation performance, reaching $29.79$ PSNR and $3.37$ gFID at $512\times512$ resolution with $32\times$ compression.
Text-to-image experiments further demonstrate its applicability beyond ImageNet: a $1.6$B-parameter DiT using DC-SAE achieves $0.84$ on GenEval and $86.007$ on DPG-Bench at $1024\times1024$ resolution with CFG.

\bibliographystyle{plainnat}
\setlength{\bibhang}{0pt}
\setlength{\bibindent}{0pt}
\bibliography{references_arxiv}

\clearpage
\appendix
\section{Appendix}

\subsection{Limitations and Future Work.}
Our evaluation covers class-conditional ImageNet generation and text-to-image generation at $1024\times1024$ resolution. However, the text-to-image study uses one $1.6$B DiT configuration and a subset of internal data together with LAION-COCO; broader validation across training scales, datasets, and generator architectures remains future work.
In addition, while we observe that semantic latents can accelerate diffusion training and that input-level pre-merge compression outperforms post-hoc token merging, we do not yet have a complete understanding of which properties of semantic encoders are most important for generation.
A deeper analysis of semantic encoder selection, as well as a principled explanation of when pre-merge or post-merge compression is preferable, remains an important direction for future work.

\subsection{Tokenizer Training Details}
\label{app:tokenizer_details}

\paragraph{Datasets.}
We use ImageNet-1K for tokenizer training.
Most experiments are conducted at $256\times256$ resolution.
For $512\times512$ generation, we initialize the tokenizer from the trained $256\times256$ checkpoint and further finetune it on $512\times512$ images.

\paragraph{Tokenizer Architecture.}
Our tokenizer contains a frozen semantic encoder, a trainable pixel encoder, a fusion projection, and a ViT decoder with Spatial DeMerger.
The semantic encoder is always frozen.
The pixel encoder, fusion projection, Spatial DeMerger, and decoder are trained for reconstruction.
The decoder architecture and discriminator largely follow RAE~\cite{zheng2025rae}.

\paragraph{Training Configuration.}
We follow the basic tokenizer training setting of RAE~\cite{zheng2025rae}, including the decoder and discriminator training setup.
We set the L1 loss weight to $\lambda_1 =1$,LPIPS loss weight to $\lambda_{\mathrm{lpips}}=1$ and the adversarial loss weight to $\lambda_{\mathrm{GAN}}=0.5$.
At $256\times256$ resolution, we train the tokenizer for $30$ epochs with a batch size of $512$.
For $512\times512$ resolution, we initialize from the converged $256\times256$ checkpoint and continue training for $20$ epochs with a batch size of $512$.

  \subsection{Throughput Comparison with DC-AE}

  As shown in Tab.~\ref{tab:throughput_comparison}, we compare the
  end-to-end throughput of DC-AE and DC-SAE under the same settings.
  All throughput numbers are measured with a batch size of 32 on a single
  NVIDIA H200 GPU using bf16, 32 batch-sizes with 10 warmup iterations and 50 measurement iterations. DC-SAE consistently
  achieves higher end-to-end throughput than the corresponding DC-AE
  baseline across all evaluated resolutions, with the speedup remaining
  above $4\times$ at every resolution.

  \begin{table}[htbp]
  \centering
  \caption{End-to-end throughput comparison between DC-AE and DC-SAE
  (imgs/s).}
  \label{tab:throughput_comparison}
  \begin{tabular}{lccc}
  \toprule
  \textbf{Model} & \textbf{$256\times256$} & \textbf{$512\times512$} & \textbf{$1024\times1024$} \\
  \midrule
  DC-AE-f32c32 & 265.6 & 69.7 & 17.4 \\
  DC-SAE-f32c64  & \textbf{1233.5} & \textbf{330.8} & \textbf{76.8} \\
  DC-SAE-f32c256 & 1054.5 & 284.4 & 67.9 \\
  \bottomrule
  \end{tabular}
  \end{table}

\subsection{Additional Analysis of Spatial DeMerger}
\label{sec:appendix_spatial_demerger_analysis}

To further demonstrate the effectiveness of Spatial DeMerger, we conduct additional experiments with QwenViT-based semantic encoders.
In particular, we consider two common vision-encoder configurations used in MLLM settings.
The first one uses the original frozen QwenViT with its native merger module.
This follows the design used in Qwen-VL, where the vision encoder consists of a $16\times$ patch-based QwenViT followed by a $2\times2$ merger trained together with the MLLM.
The second one replaces the original QwenViT merger with a simple $2\times2$ average pooling operation, which is another commonly used way to reduce the spatial resolution of vision tokens in MLLM vision encoders.

In all experiments, the semantic encoder is kept fully frozen, and no additional pixel encoder is used, unlike the settings studied in the main paper.
This setup isolates the role of Spatial DeMerger under a stricter reconstruction setting, where the decoder must recover spatial details only from the frozen semantic representation.
As shown in Tab.~\ref{tab:appendix_spatial_demerger_qwenvit}, Spatial DeMerger consistently improves reconstruction quality for both the original QwenViT merger and the $2\times2$ average pooling variant.
These results suggest that Spatial DeMerger is especially useful for RAE-like settings, where the encoder is fully frozen, and the decoder receives only sparse semantic tokens without auxiliary pixel-level features.
\begin{table}[htbp]
\centering
\scriptsize
\caption{
Additional analysis of Spatial DeMerger with QwenViT-based semantic encoders.
We compare the original QwenViT merger and a $2\times2$ average pooling variant,
with and without Spatial DeMerger.
Both semantic encoders are frozen, and no additional pixel encoder is used.
}
\label{tab:appendix_spatial_demerger_qwenvit}
\setlength{\tabcolsep}{3pt}
\resizebox{0.6\linewidth}{!}{%
\begin{tabular}{lcc}
\toprule
\multirow{2}{*}{Semantic encoder setting} 
& \multicolumn{2}{c}{Spatial DeMerger} \\
\cmidrule(lr){2-3}
& w/o DeMerger & w/ DeMerger \\
\midrule
QwenViT w/ original merger 
& 15.39 & 17.52 \\
QwenViT w/ $2\times2$ avg pooling 
& 15.50 & 17.64 \\
\bottomrule
\end{tabular}
}
\end{table}

\end{document}